\documentclass{article}

\PassOptionsToPackage{numbers, compress}{natbib}
\usepackage[final]{neurips_2020}
\usepackage[utf8]{inputenc} 
\usepackage[T1]{fontenc}    
\usepackage{hyperref}       
\usepackage{url}            
\usepackage{booktabs}       
\usepackage{amsfonts}       
\usepackage{nicefrac}       
\usepackage{microtype}      
\usepackage[algo2e, noend, noline, ruled, linesnumbered]{algorithm2e}
\usepackage[colorinlistoftodos,prependcaption,textsize=tiny]{todonotes}
\usepackage{amsmath}

\title{Human-Agent Cooperation in Bridge Bidding}

\author{
Edward Lockhart \\ \texttt{locked@google.com} \\ DeepMind \and
Neil Burch \\ DeepMind  \and
Nolan Bard \\ DeepMind  \and
Sebastian Borgeaud \\ DeepMind  \and
Tom Eccles \\ DeepMind  \and
Lucas Smaira \\ DeepMind  \and
Ray Smith \\ DeepMind  
}

\begin{document}

\maketitle

\begin{abstract}
  We introduce a human-compatible reinforcement-learning approach to a cooperative game,
  making use of a third-party hand-coded human-compatible bot to generate initial training data
  and to perform initial evaluation.
  Our learning approach consists of imitation learning, search, and policy iteration. 
  Our trained agents achieve a new state-of-the-art for bridge bidding in three settings: an
  agent playing in partnership with
  a copy of itself; an agent partnering a pre-existing bot; and an agent partnering a human player.
\end{abstract}

\section{Introduction}

Deep reinforcement learning has had great success in two-player, zero-sum games.
Key to the success of the algorithms used for these games is the fact that if neither player can unilaterally improve, the policies must be globally optimal \cite{minimax}, and therefore considering only unilateral policy deviations is sufficient to achieve optimal play.
On the other hand, collaborative multi-agent environments have many local optima -- agents must coordinate their strategies with each other to perform well.
We are particularly interested in human-compatible agents for such domains, because
of the general importance of agents that can cooperate with humans.

A type of environment where the need for coordination is particularly stark is a communication game; if one agent unilaterally diverges from an agreed ``language'', the results are likely to be very poor.
The bidding phase of bridge is one such domain, involving grounded communication and competition between two teams of two. It has been extensively studied by human players and has coordination and cooperation with a partner as key features. There are many fundamentally different equilibria in this domain; in particular, humans have devised many mutually-incompatible communication systems.

We address the problem of training an agent to learn a human-compatible policy for bridge bidding. Using imitation learning, search, and policy iteration we train an agent to cooperate with a strong hand-coded human-compatible bot (WBridge5), and evaluate it in play with a human expert.

Our approach starts by imitation-learning from a dataset of deals played by WBridge5 to obtain an initial human-compatible policy. We then use a search algorithm that aims to improve a given policy, combining the prior policy with the results of rollouts. Starting with our learned model of WBridge5, we perform several (up to 16) rounds of policy iteration. In each round we first generate a dataset using the policy and search, and then update the policy from this dataset. If we are aiming for human-compatibility, we use the imitation-learned model as a partner; if we are aiming for maximum team performance, we use ourself as our partner. Finally, at test time we optionally apply a further larger search to improve our policy.

\section{Background}

\subsection{Bridge Bidding}

Contract bridge is a zero-sum game played between two teams of two players. We summarize key aspects of the rules here; see the Laws of Duplicate Bridge~\cite{wbflaws} for more detail.
The four players are conventionally labelled North, East, South, and West, and the two partnerships are North-South and East-West. Each player is dealt a private hand of 13 cards. The game then proceeds in two phases: first \textit{bidding}, then \textit{play}. We describe the play first, since this grounds the bidding.

In the play phase, 13 \textit{tricks} are played. The first player to a trick may play any of their remaining cards. Subsequent players must play a card of the same suit as the first player if they have any, or else any card from their hand. The trick is won by the highest \textit{trump} if any were played, or otherwise the highest card of the suit led. Scoring is based on the division of tricks between the two sides.

In the bidding phase, players take turns to make a \textit{call} which is one of Pass, Double, Redouble, or a \textit{bid}. Bids specify a trick target and a trump suit (or no trump). The collective sequence of bids made must be ascending, with bids ordered first by number of tricks and then by trump suit. The final bid, or \textit{contract} sets the \textit{declarer}, trump suit and trick target for the play phase.

At the end of the play, the declarer's side receives a positive score if they met or exceeded the trick target; otherwise their score is negative. Higher contracts receive significant bonuses if made; this incentivizes players to use low-level bids conventionally to exchange information and determine whether a high contract is feasible~\footnote{These conventions may be extremely intricate in expert play, for example enabling one player to ask their partner whether they have a single specific card. Some partnerships have hundreds of pages of detailed notes.}. Conversely, a player with a weak hand may make a high-level bid to restrict the ability of the opponents to communicate in this way. Bids are thus grounded communication actions, serving multiple purposes. Each bid communicates information about the bidder’s cards, and restricts the later bids of other players, as well as being a possible contract.

The rules require that players be fully informed of the bidding system used by their opponents, both so that they are able to draw the correct inferences from the actions taken and so that they are able to adapt their subsequent system. In tournament
play, this is achieved through a combination of disclosing agreements to the opponents in advance (allowing preparation), disclosing agreements during the course of the deal (at which point no discussion is permitted), and restricting the nature of permissible agreements, hence reducing the adaptations that may be necessary.

Following previous work \cite{bridge-pql, bridge-dnn, bridge-simple, jps} 
we consider only the bidding phase of the game; after the bidding, we assign rewards based on \textit{double-dummy play}; that is, the result of the play phase computed assuming perfect play for all players given perfect information. This bidding-only game is a slightly different game from bidding in the full game of bridge for two reasons. Firstly, because there is a cost during the play phase from having revealed information during the bidding phase, which may affect the best bidding strategy. Secondly, because the expected result of the imperfect-information card play phase may be different from the result of optimal perfect-information play. However, the average score will be close, and a statistical study by a world-class player concluded that ``actual [expert-level] play versus double-dummy is pretty close, so analyses based on double-dummy play should generally be on track''\cite{rp8j45}.

\subsection{WBridge5}

WBridge5 \cite{wb5} by Yves Costel is a strong computer bridge program which has won the World Computer Bridge Championship six times, most recently in 2018, and has been runner up seven times
\cite{computerbridge}. The format of this competition is the same as our team play setting.
WBridge5 is the standard comparator for machine-learning agents because it is free software,
although not open source. Some information on the algorithms used by the program may be gleaned from \cite{boosting}.

WBridge5 can be programmatically interacted with using the Blue Chip Bridge protocol \cite{bluechip}.
For training purposes, we used a public dataset of one million deals played by WBridge5 \cite{wb5data}. In this dataset, WBridge5 is configured to play according to the Standard American Yellow Card system \cite{sayc}, which was devised by the American Contract Bridge League in the 1980s and has become popular in online play as a simple system for casual partnerships, which makes it a good starting point for human-agent cooperation.

\subsection{Tasks}\label{sec:tasks}

We consider two distinct tasks: human-compatible and partnership play.

For human-compatible play, the aim is to produce agents which can play well in a partnership with humans. Since WBridge5 is designed to play a human-like strategy, we use playing with WBridge5 as a proxy for training and evaluation of this task. We then evaluate the best agent in this setting as a partner to a human expert.

For partnership play, the aim is produce agents that can play well in a partnership with themselves. This is the task which has been addressed in previous work
     \cite{bridge-pql, bridge-dnn, bridge-simple, jps}, with the usual evaluation metric being performance against WBridge5.
     
In line with prior work, we ignore the system restrictions and disclosure requirements of bridge. However evaluating our performance as a partner to WBridge5 implicitly ties us to the Standard American Yellow Card system, which does meet these requirements.

\subsection{Prior Work}\label{sec:prior}

Ginsberg \cite{gib} introduced
the Borel search algorithm to improve a bridge bidding policy expressed as a set of rules in a domain-specific language:
\begin{quote}
To select a bid from a candidate set $B$, given a database $Z$ that suggests
bids in various situations:
\begin{enumerate}
    \item Construct a set $D$ of deals consistent with the bidding thus far.
    \item For each bid $b \in B$ and each deal $d \in D$, use the
database $Z$ to project how the auction will continue
if the bid $b$ is made. (If no bid is suggested by the
database, the player in question is assumed to pass.)
Compute the double dummy result of the eventual
contract, denoting it $s(b,d)$.
\item Return that $b$ for which $s(b,d)$ is maximal. 
\end{enumerate}
\end{quote}

Ginsberg used Borel search in the bridge-playing program GIB at test-time
to potentially override the action selection given by the rules. 
The paper describes a
number of heuristics to attempt to avoid failures when there was a weakness in the original
policy. We did not find these to be necessary in our approach, we believe both because our starting
policy was stronger, and also because our policy iteration would eventually remove such weaknesses.
Ginsberg mentions two specific classes of problems. First:

\begin{quote}
Suppose that the database $Z$ is somewhat conservative
in its actions. The projection in step 2 leads each player
to assume his partner bids conservatively, and therefore
to bid somewhat aggressively to compensate. The partnership
as a whole ends up overcompensating.
\end{quote}

In our partnership policy iteration approach, both our
policy and our partner's policy will gradually get more aggressive, and the iteration should converge
at a locally optimal level of aggression. Second:

\begin{quote}
Worse still, suppose that there is an omission of some
kind in $Z$; perhaps every time someone bids $7\diamondsuit$ the
database suggests a foolish action. Since $7\diamondsuit$ is a rare
bid, a bidding system that matches its bids directly to
the database will encounter this problem infrequently. 

GIB, however, will be much more aggressive, bidding $7\diamondsuit$
often on the grounds that doing so will cause the
opponents to make a mistake. In practice, of course, the
bug in the database is unlikely to be replicated in the
opponents' minds, and GIB's attempts to exploit the gap
will be unrewarded or worse. 
\end{quote}

Our policy iteration approaches use our learned policy as an opponent model, so that if the starting policy does
have a weakness like this, it will be repeatedly triggered in training data; subsequent search should find
better countermeasures to the $7\diamondsuit$ bid, which will be incorporated into the learned policy, eventually making
the $7\diamondsuit$ action less attractive.

While the proposed combination of imitation learning, search, and policy iteration is novel, very similar components have been used in past successful game-playing programs.

Combining a deep neural network policy with a search procedure at training time to
generate improved training data, and at test time to improve upon the raw neural
network policy was introduced in \cite{rlclass} and has been successfully used in a number of games, including Hex \cite{expit}, Go \cite{alphagozero}, Chess and Shogi \cite{alphazero}.
As in \cite{alphagozero}, we start the policy iteration process
by learning from a dataset of games played by strong players, with human-compatibility as an additional motivation.

Recent work has shown that standard reinforcement
learning self-play techniques can lead to good performance on bridge bidding \cite{bridge-simple}. Following this work, we use standard neural network architectures without any special auxiliary losses.

Improving a prior policy using search has achieved state-of-the-art results in the fully-cooperative
game of Hanabi~\cite{sparta}. We extend this approach by sampling possible world states (as considering them all would be infeasible~\footnote{There are $6e18$ distinct possible distributions of the hidden cards from the point of view of a searching player. For comparison, the search algorithm in \cite{sparta} is only used when there are fewer than $1e4$ possible states.}), and by embedding the search in a policy iteration loop.

Other recent work examined approaches for exploring the policy space in cooperative games including bridge bidding \cite{jps}.
 Our work does not require this exploration, as we start from a fairly strong policy.

Improving upon a policy while maintaining the ability to cooperate with a human, or
a bot as a proxy for a human, is also relevant in the natural language domain, where
there is recent work on preventing ``language drift''.
In  \cite{dealornodeal}, episodes of supervised learning from human demonstrations are interleaved with self-play to reduce divergence from the human policy. In  \cite{seeded}, a single student learns from data generated by successive generations of fine-tuned teachers. One approach discussed in \cite{language-drift} is using rewards from partnering with a fixed listener to prevent drift in the policy of a speaker, which is analogous to our use of an imitation-learned
model as a partner.

\section{Methods}

\subsection{Imitation learning}\label{sec:imitation_learning}

The first stage of our method is to learn a model which can predict the bids made by WBridge5. We learn a policy network $P_\Phi(a|f)$ which gives probabilities for each action $a$ given features $f$ describing the state of the game. 
We use the OpenSpiel implementation of bridge  \cite{openspiel} to track the game state and generate a 480-element input feature vector as follows:

\begin{tabular}{lr}
Phase of the game (always the bidding phase) & 2\\
Vulnerability (always neither side vulnerable) & 2 \\
Per player, did this player pass before the opening bid? & 4 \\
Per player and bid, did this player make this bid? & 140 \\
Per player and bid, did this player double this bid? & 140 \\
Per player and bid, did this player redouble this bid? & 140 \\
13-hot vector indicating the cards we hold & 52 \\
\end{tabular}

This representation has perfect recall, i.e. it is possible to reconstruct the entire sequence of actions and observations from the current observation.

Our policy network is a 4-layer MLP with 1024 neurons per layer, ReLU non-linearities, and a softmax output, giving probabilities for each of the 38 actions (35 bids, Pass, Double, and Redouble).

We train on
a dataset of WBridge5 self-play \cite{wb5data}, selecting decision points uniformly at random, with a minibatch size of 16.
We used 200,000 steps of the Adam optimizer with learning rate 0.0003 to optimize a cross-entropy loss; we ran experiments with various auxiliary losses, such as predicting the cards in partner's hand, but found that these did not improve performance.

\subsection{Search}\label{sec:search}

We extend the Borel search discussed in \ref{sec:prior} to use neural policies and a soft update.
We use the term \textit{particle} for a single determinization of the hidden information, i.e. a complete deal which is consistent with the information possessed by one player. Our extended Borel search, shown in Algorithm~\ref{alg:borel}, uses the rollout evaluation in Algorithm~\ref{alg:rollout} for several particles to evaluate a small candidate set of actions.

There are several differences between Ginsberg's Borel search and our search algorithm. Firstly, we represent policies by neural nets, which may be different for the different players. We then select candidate particles at random, based on our estimated probability that they would result in the actual actions observed. When performing rollouts, we select actions according to a stochastic policy, allowing for more diversity of possible outcomes.
Finally, our prior policy is adjusted towards the search results, not replaced by them. 

The policies used to filter particles and generate rollouts are approximations when partnering WBridge5 or a human, because we don't have access to the true policy in these cases. We optionally use a search to improve the policy at test time, which results in a divergence between the policy we follow and the policy we use to filter particles and perform rollouts. It is infeasible to use the search-augmented policy here because to do so would require search-within-search, when the search process is already expensive.

\begin{algorithm2e}[!ht]
\SetKwInOut{Input}{input}
\SetKwInOut{Output}{output}
\Input{$h$ --- public history so far}
\Input{$a$ --- action to evaluate}
\Input{$ \left\{ p_i \right\} $ --- private information for each player $i$}
\Input{$ \left\{ \pi^i \right\} $ --- rollout policy for each player $i$}
\caption{Rollout Value \label{alg:rollout}}
$h' \leftarrow h + a$ \\
\While{$h'$ is not terminal}{
  $j \leftarrow \operatorname{ActingPlayer}(h')$ \\
  sample action $a'$ from $\pi^j(h', p_j)$ \\
  $h' \leftarrow h' + a'$ \\
}
\Return reward for $\operatorname{ActingPlayer}(h)$ of the final contract $h'$, using double-dummy analysis
\end{algorithm2e}

\begin{algorithm2e}[!ht]
\SetKwInOut{Input}{input}
\SetKwInOut{Output}{output}
\SetKwInOut{Parameter}{parameter}
\Parameter{$t$ --- temperature}
\Parameter{$R_{min}$ --- minimum rollouts to use the result of the search}
\Parameter{$R_{max}$ --- maximum rollouts for the search}
\Parameter{$P_{max}$ --- maximum particles to consider}
\Parameter{$k$ --- maximum actions to consider}
\Parameter{$p_{min}$ --- minimum action probability to consider}
\Input{$h$ --- public history so far}
\Input{$p_s$ --- private information for the searching player $s$}
\Input{$\pi_{prior}$ --- prior policy for the searching player}
\Input{$\left\{\pi^i\right\}$ --- rollout policies for each player $i$}
\Output{$\pi_{posterior}$ --- posterior policy for the searching player}
\caption{Borel Search with Non-Deterministic Model \label{alg:borel}}
$s \leftarrow \operatorname{ActingPlayer}(h)$ \\
$A \leftarrow$ top-$k$ actions from $\pi_{prior}$ with probability at least $p_{min}$\\
$V(a) \leftarrow 0$ for $a \in A$\\
$R \leftarrow 0$\\
$P \leftarrow 0$\\
\While{$P < P_{max}$ and $R < R_{max}$}{
  jointly sample private information $\left\{p_{-s}\right\}$ for players other than $s$ uniformly, consistent with $p_s$ \\

  form the particle $\left\{ p_i \right\} = \left\{ p_s \right\} \cup \left\{ p_{-s} \right\}$ \\
  $P \leftarrow P + 1$ \\
  \For{ each $h'a' \sqsubseteq h$}{
    $j \leftarrow \operatorname{ActingPlayer}(h')$ \\
    with probability $1 - \pi^{j}(a' | h', p_{j})$, skip to next particle
    }
  \For{$a \in A$}{
    $V(a) \leftarrow V(a) + \operatorname{RolloutValue}(h, a, \left\{ p_i \right\}, \left\{ \pi^i \right\} )$ \\
    }
    $R \leftarrow R + 1$\\
    }
\If{$R > R_{min}$}{
   $\pi_{posterior} \propto \pi_{prior} \times \exp \left( \frac{V(a)}{t\sqrt{R}} \right)$\\}
\Else{
    $\pi_{posterior} = \pi_{prior}$\\}
\end{algorithm2e}

\subsection{Policy Iteration}
We implement a policy iteration loop by generating experience from a search policy as set out in \ref{sec:search}. We denote the current reinforcement learning policy $\pi_l$; it is initially set to the WBridge5 imitation policy $\pi_b$, as learned in \ref{sec:imitation_learning}, using the same neural network architecture. At each iteration, we learn a new updated policy $\pi_{l'}$ by imitating the policy produced by the search. The search configuration is slightly different depending on the aim of the learning task (that is, learning to play with an existing agent or in self-play).
\begin{itemize}
    \item For both tasks, we use $\pi_l$ for the prior policy, and to rollout our own and the opponents' actions during search.
    \item For learning to play a policy compatible with WBridge5, we use $\pi_b$ for our partner's policy.
    \item For learning to play in partnership with ourselves, we use $\pi_l$ for our partner's policy.
\end{itemize}

Note that this choice of opponent policies is slightly counter-intuitive: in our evaluations, the opponent agents are WBridge5, so the imitation policy $\pi_b$ might be a better model than $\pi_l$. Empirically, we found that using $\pi_b$ in search produces policies which perform well against $\pi_b$, but not against WBridge5. We conclude that the search policy exploits weaknesses in $\pi_b$ which are not present in WBridge5 and which the policy iteration loop is able to remove.

Finally, we perform a search at test time using a larger search budget, to improve our policy further. For this search, we use greedy versions of the policies for rollouts, selecting the highest-probability action for each player.

The search parameters during policy iteration and at test time were as follows:

\begin{center}
\centering
\begin{tabular}{|l|r|r|}
\hline
Parameter & Policy Iteration & Test Time \\
\hline
$R_{min}$ & 1 & 100 \\
$R_{max}$ & 30 & 1,000 \\
$P_{max}$ & 100,000 & 100,000 \\
$t$ & 300 & 100 \\
$k$ & 4 & 4 \\
$p_{min}$ & 10$^{-4}$ & 10$^{-4}$ \\
\hline
\end{tabular}
\end{center}

The acting network was updated every hour, which is approximately every 10 million learner observations. The experiments were stopped when performance of $\pi_l$ against $\pi_b$ appeared to be levelling off; this was after 6 network updates and 50 million learner observations for compatible learning, and 16 network updates and 170 million learner observations for partnership learning.

\section{Results}

\subsection{Imitation learning}

Our imitation model achieved 93.9\% accuracy in predicting the actions of WBridge5 on a held-out test set.

\subsection{Bot evaluation}

We evaluate each of our agents on the two tasks described in \ref{sec:tasks}.
In both cases, we use IMPs\cite[Law~78B]{wbflaws} to rescale raw score differences; this is consistent with prior work and is the usual practice in tournament bridge play. Let $S(pqrs)$ represent the score obtained by
the North-South partnership when the deal is played with agent $p$ as North, agent $q$ as East, agent $r$ as South, and agent $s$ as West; and let $w$ represent WBridge5 and $a$ the agent under evaluation.

To evaluate WBridge5-compatible play, we compare deals played by four WBridge5 bots to the same deals played by three WBridge5 bots and one of the agent being tested. Specifically, our metric for a single deal is:
\begin{align*}
    &\operatorname{IMP}\left(S(awww) - S(wwww)\right) 
    - \operatorname{IMP}\left(S(waww) - S(wwww)\right) \\
    + &\operatorname{IMP}\left(S(wwaw) - S(wwww)\right) 
    - \operatorname{IMP}\left(S(wwwa) - S(wwww)\right)
\end{align*}
 This is positive for agents which are better partners to WBridge5 than WBridge5 itself. 

To evaluate team play, each deal is played twice, once with our agent playing the North-South hands, and once with our agent playing the East-West hands.
Specifically, our metric for a single deal is
$ \operatorname{IMP}\left(S(awaw) - S(wawa)\right) $.
This is positive if the agent achieves better results than WBridge5.

In both cases, we express our results in average IMPs per deal over a set of 10,000 held-out deals.
This is considerably more evaluation deals than prior work, which accounts for at least some of our tighter error bounds.

\begin{table}
\centering
\begin{tabular}{|c|c|c|}
    \hline
    Algorithm & WBridge5-compatible & Team \\
    \hline
    Penetrative Q-Learning \cite{bridge-pql} & n/a & $+0.20$ \\
    Competitive Bridge Bidding with DNNs \cite{bridge-dnn} & n/a & $+0.25$ \\
    Simple is Better \cite{bridge-simple} & n/a & $+0.41 \pm 0.27$ \\
    Joint Policy Search \cite{jps} & n/a & $+0.63 \pm 0.22$ \\
    Imitation Learning & $-0.12 \pm 0.05$ & $-0.11 \pm 0.04$ \\
    
    Compatible Policy Iteration (network only) & $+0.28 \pm 0.06$ & $+0.36 \pm 0.05$ \\
    
    Compatible Policy Iteration + test-time search & $\mathbf{+0.48 \pm 0.06}$ & $+0.56 \pm 0.05$ \\
    
    Partnership Policy Iteration (network only) & $+0.11 \pm 0.06$ & $+0.57 \pm 0.05$ \\
    
    Partnership Policy Iteration + test-time search & $+0.12 \pm 0.07$ & $\mathbf{+0.85 \pm 0.05}$ \\
    
    \hline
\end{tabular}
\caption{IMPs per deal for learned agents. The errors shown are the standard error of the mean.}
\end{table}

As we expect, the agents which learn to partner WBridge5 and agents which learn to partner themselves each do better in their respective evaluation settings. Also as expected, test-time search improves performance, especially where the rollout policy $\pi^i$ used for our partner in the search aligns closely with the player we are partnering.

We note that for teams play, there are important differences between our algorithm and previous work which affect this comparison.
\begin{itemize}
\item Our agents are initialized to a similar policy to WBridge5, and remain somewhat similar throughout training, as evidenced by the fact that all the agents from policy iteration are better partners to WBridge5 than another copy of WBridge5. In previous work, learning has been without reference to human policies, and so this does not hold. This means our agents are less likely to gain points by ``confusing'' WBridge5 with unexpected conventions (which would be illegal in the full game, where conventions must be disclosed
to the opponents either in advance or during the course of play).

\item Our agents have explicitly trained to beat a model of WBridge5's play, whereas previous work has not. It is possible that the agent is exploiting weaknesses in WBridge5 rather than exhibiting strong play itself. Note however that any weakness would have to be present both in WBridge5 and also in our model of it, but not be removed by our policy iteration loop.

As a partial test of this, we evaluate our agent's performance on the subset of deals where one of the two partnerships does not make a bid with any configuration of agents.
Roughly 20\% of deals fall into this category. We can be sure that performance on these deals does not arise from exploiting weaknesses in WBridge5's policy, although there is generally less scope for interesting bidding in these cases, so we should expect relative scores to be smaller. Our best scores on this subset of deals are +0.37 IMPs/deal in the partnership setting and +0.24 IMPs/deal in the compatible setting. This confirms that our agents are genuinely improving on WBridge5 and not solely exploiting its weaknesses.

\item Duplicate bridge has four possible scoring tables, known as vulnerabilities \cite[Law~77]{wbflaws}. The deals in the WBridge5 dataset were all played with ``neither side vulnerable'' in order to allow statistical analysis across the whole dataset. We therefore used this single scoring table throughout our experiments. Since this vulnerability gives the smallest absolute scores, this choice is likely to have reduced our reported performance compared to prior work which uses a mixture of all four scoring tables.
\end{itemize}

\subsection{Human Evaluation}

We take the final network from compatible policy iteration and evaluate it as a partner to a human expert (one
of the authors), with no test time search.
The human plays each deal eight times, once in each seat with WBridge5 as a partner, and once in each seat with our agent as a partner;
the opponents are WBridge5 in both cases. The order in which the human partners the bots is randomized
for each deal independently so that the human does not know at the start of the deal which of the two possible bots they
are partnering. Representing the human as $h$, our scoring metric is:
\begin{align*}
    &\operatorname{IMP}\left(S(hwaw) - S(hwww)\right) 
    - \operatorname{IMP}\left(S(whwa) - S(whww)\right) \\
    + &\operatorname{IMP}\left(S(awhw) - S(wwhw)\right) 
    - \operatorname{IMP}\left(S(wawh) - S(wwwh)\right)
\end{align*}

Instructions for the human expert were as follows:
\begin{itemize}
    \item Bid each hand according to Standard American Yellow Card, selecting the calls you would make with an unfamiliar expert human partner playing the same system.
    \item Do not take advantage of information gained from earlier plays of the same deal.
    \item The second time you play a deal in a particular seat, make the same call as the first time if the situation is identical. If the situation is merely similar, be consistent with the previous play-through unless there is a clear reason to judge differently.  
\end{itemize}
This is a slightly artificial setting, but allows a meaningful comparison with relatively few
deals played.
Over an evaluation set of 32 deals, the agent outperforms WBridge5 by 0.97 IMPs/deal, with standard error of the mean 0.76.

Qualitative observations from the human expert, reviewing the hands after play:
\begin{itemize}
    \item Our agent conforms to the Standard American Yellow Card system; the differences to WBridge5 are of judgement rather than system.
    \item Our agent is slightly more aggressive in competitive auctions, matching modern expert practice.
    \item Our agent generally prefers simpler, more direct auctions, which are more robust to slight differences in interpretation.
    \item Little, if any, of the agent's improved performance was due to differences from the full game of bridge; i.e. the agent did not gain points because WBridge5 was concealing information in anticipation of the play phase, nor because of the double-dummy play assumption.
\end{itemize}

\section{Summary}

We introduced the problem of learning human-compatible policies for bridge bidding. This is an interesting task to tackle because it requires learning to communicate and collaborate with a human in a challenging domain -- one which is played competitively and has been extensively studied.
Using a combination of imitation learning, policy iteration and search, we trained agents to improve on a hand-coded bot playing a human-like strategy. Our agents improved on this baseline bot, both when playing with the bot itself and also with an expert human. We believe that the strong collaborative performance is owing to our combination of a good starting policy and searching with the assumption that partner sticks to that starting policy, thereby penalizing incompatible policy innovations.

Our partnership learning approach achieves new state of the art performance when playing as a pair against WBridge5, which is the problem addressed in previous work. Perhaps surprisingly, these agents also maintain compatibility with WBridge5, performing better as a partner than WBridge5 itself does. We believe this compatibility is owing to the strong starting policy and the soft policy updates, which will result in conservative policy exploration rather than finding a radically different local optimum.

\section*{Broader Impact}

Agents that cooperate with humans are an important long-term goal of AI research, with the potential for significant societal benefit.
Approaches along the lines that we describe here could eventually be used in a wide range of environments to improve on existing cooperative agents without losing human-compatibility, or to learn an initial policy by imitation learning from human data and then improve on it, again without losing human-compatibility. Humans could in turn learn from the improvements that the agent finds.

Our agent learns to collaborate only with one particular human policy.
The requirement for an existing agent or a large corpus of data on which to train means that our approach may not be readily extensible to conventions used by minority groups where this prior work or large datasets do not exist.
On the other hand, the possibility to generate a high-quality human-compatible agent through demonstration and self-play may be more feasible than hand-engineering for such communities.

Our imitation learning approach requires a homogeneous dataset, in which the same conventions are used throughout. This means our approach is not directly applicable to situations where a group of people with diverse behaviour interact.

\end{document}